\documentclass[utf8]{frontiersSCNS} 

\usepackage{url,hyperref,microtype,subcaption}
\usepackage[onehalfspacing]{setspace}

\def\keyFont{\fontsize{8}{11}\helveticabold }
\def\firstAuthorLast{Sinapayen {et~al.}} 
\def\Authors{Lana Sinapayen\,$^{1,2}$, Atsushi Masumori\,$^{3,*}$ and Takashi Ikegami\,$^{3}$}
\def\Address{
$^{1}$Sony Computer Science Laboratories, Inc., Shinagawa-ku, Tokyo, Japan\\
$^{2}$Earth-Life Science Institute, Tokyo Institute of Technology, Tokyo, Japan\\
$^{3}$Ikegami Laboratory, Department of General Systems Studies, Graduate School of Arts and Sciences, The University of Tokyo, Tokyo, Japan
}
\def\corrAuthor{Room 324B Building 16 3-8-1 Komaba, Meguro, Tokyo 153-0041 Japan}

\def\corrEmail{masumori@sacral.c.u-tokyo.ac.jp}

\graphicspath{{figures/}}

\begin{document}
\onecolumn
\firstpage{1}

\title[Reactive, Proactive, and Inductive Agents]{Reactive, Proactive, and Inductive Agents: An evolutionary path for biological and artificial spiking networks} 

\author[\firstAuthorLast ]{\Authors} 
\address{} 
\correspondance{} 

\extraAuth{}

\maketitle

\begin{abstract}

\section{}
Complex environments provide structured yet variable sensory inputs.
To best exploit information from these environments, organisms must evolve the ability to anticipate consequences of unknown stimuli, and act on these predictions. We propose an evolutionary path for neural networks, leading an organism from reactive behavior to simple proactive behavior and from simple proactive behavior to induction-based behavior.
Through in-vitro and in-silico experiments, we define the conditions necessary in a network with spike-timing dependent plasticity for the organism to go from reactive to proactive behavior. 
Our results support the existence of specific evolutionary steps and four  conditions necessary for embodied neural networks to evolve predictive and inductive abilities from an initial reactive strategy. We extend these conditions to more general structures.

\tiny
 \keyFont{ \section{Keywords:} Neural network, Spiking neural network, Predictive coding, LSA, STDP} 
\end{abstract}

\section{Introduction}
\label{sec:intro}
There are surprisingly few hypotheses about how cognitive functions such as generalization and prediction might have evolved. The ability to generate predictions especially, is often assumed to be a given in evolutionary simulations.
Dennett proposes an evolutionary path by dividing living systems in four classes (\cite{Dennett1995}): Darwinian creatures, with hard-wired reactions acquired through evolutionary processes; Skinnerian creatures, with phenotypic plasticity to acquire suitable sensory-motor coupling in their environment; Popperian creatures, which can predict the outcome of their actions; and Gregorian creatures, which use knowledge acquired from their predecessors. Dennett also proposes that the biological creatures must have evolved in this order.
The three types of agents discussed in this paper have partial overlap with Denett's classification. We focus on a specific definition of ``agent": a neural network embedded in a body, and able to perform actions that cause changes in the environment. Note that we only consider agents that are able to \textbf{learn during their lifetime}. Our three types of agents are as follows:
\begin{itemize}
    \item Reactive agents learn during their lifetime how to react to environmental stimuli. These agents correspond to Skinnerian creatures.
    \item Through evolution, if reactive agents become able to act in anticipation of a stimulus before receiving that stimulus, we call them proactive agents: they can perform actions based on prediction. Proactive agents correspond to Popperian creatures. The idea that organisms constantly try to predict their environment has been credited for explaining typical perception (\cite{Rao1999}), illusions (\cite{Raman2016, Edwards2017, Watanabe2018}), hallucinations (\cite{Powers2016, Suzuki2017}), and even consciousness (\cite{Seth2012}).
    Without going so far, proactive agents have clear advantages over reactive agents: they can avoid or select behaviors before experiencing undesirable (resp. desirable) consequences.
    \item
    Inductive agents fit in-between Denett's Popperian and Gregorian creatures. Inductive agents are able to make generalizations about learned stimuli, and to react to new stimuli based on these generalizations. As Gregorian creatures must be able to apply knowledge from their predecessors to their own situation, we can argue that inductive agents must come before Gregorian creatures in terms of evolution.
\end{itemize}

The distinction between reaction and prediction can be unclear, because learning to react to stimuli is sometimes inseparable from learning to predict consequences of stimuli. Here, we define a prediction as the information generated inside an agent, equal to the content of an external input, but preceding that input in time. The ability to make predictions therefore implies the existence of a generative model inside the agent.

In \cite{chung2009evolution, kwon2008internal}, Chung and Kwon show through simulated evolution experiments that neural networks with predictable dynamics are better at generalizing what they learned to novel tasks. The predictability of network dynamics does not correlate with better performance on known tasks, but it does correlate with better performance on new tasks, showing that the networks have better generalization abilities.
Unfortunately these results do not tell us about the actual predictive ability of the networks. Predictable networks do better, but are the networks themselves performing any kind of predictions on the environment?

The contribution of this paper is to propose a theory of how and why predictive and generalizing abilities might have evolved in neural networks. To the authors' knowledge, there is currently no theory relating these concepts or explaining how, in practice, they would have emerged from an evolutionary point of view.
In this paper, we focus on the evolutionary transition from reactive agents to proactive agents, and from proactive agents to inductive agents.

\begin{figure}[t]
\begin{center}
\includegraphics[width=12cm]{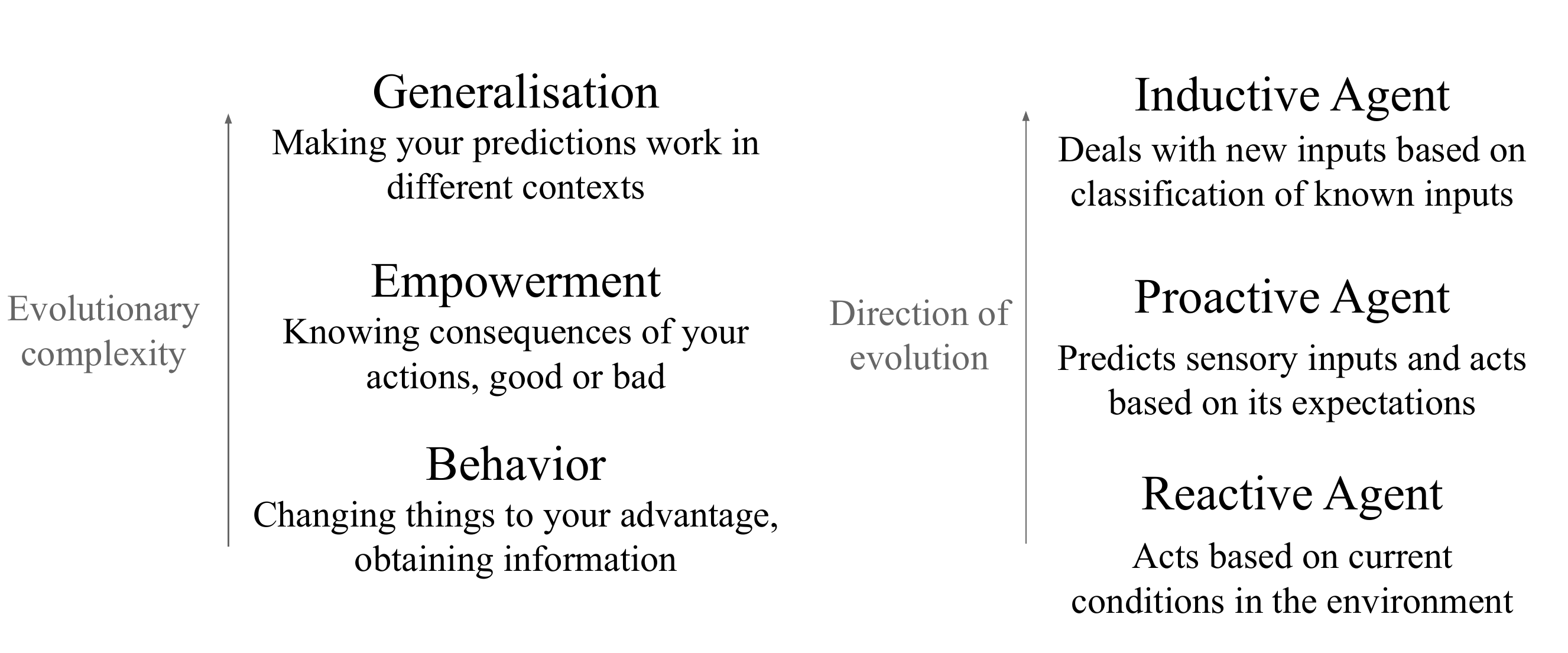}
\end{center}
\caption{Our proposal of evolutionary path for reactive agent, proactive agent, and inductive agent. An organism needs behavior; this need is met by the evolution of actuators to modify the world. It needs to know the consequences of its actions without necessarily acting them out; this need is met by evolving predictive abilities. Finally, the agent needs to generalize these predictions, which is the role of classification. Prediction emerges to improve the agent's actions, and classification emerges to improve the agent's predictions.}
\label{fig:evolution}
\end{figure}

Fig.~\ref{fig:evolution} shows how these three functions are linked, and the environmental needs these functions can fulfill for an embodied agent. An agent first needs behavior: the ability to change its environment to its advantage. This need is met by reactive agents: through action, they can change their environment.
An agent that can only react to the environment does not have much control on its future. Increasing this control is increasing empowerment. Empowerment is a quantity defined as how much the agent can potentially influence the environment. It quantifies not what the agent actually does, but what it can ``potentially'' do to influence the environment \cite{Klyubin2005}. To increase its empowerment, an agent needs to predict the consequences of its actions: these are proactive agents.
Finally comes a need for generalization: the ability to recognize new inputs as being similar to known inputs, and to generate appropriate predictions. This need is met by inductive agents through classification.
We argue that action, prediction and classification emerge from the bottom up: prediction emerges from action and classification emerges from prediction.

As a practical example of learning rule that can be used for the three functions, we use results from our experiments with Learning by Stimulation Avoidance (LSA). 
LSA (\cite{Sinapayen2016}) is a property exhibited by spiking networks coupled with Spike-Timing Dependent Plasticity rules (STDP; \cite{Caporale2008}): the networks learn the behaviors that stop external stimulation, and they learn to avoid the behaviors that start external stimulation. 
Neither LSA nor STDP are considered as necessary mechanisms in this paper; we take them as one practical example of how our ideas can be implemented, and we acknowledge that other implementations are possible.

\begin{figure}[t]
\centering
\includegraphics[width=15cm]{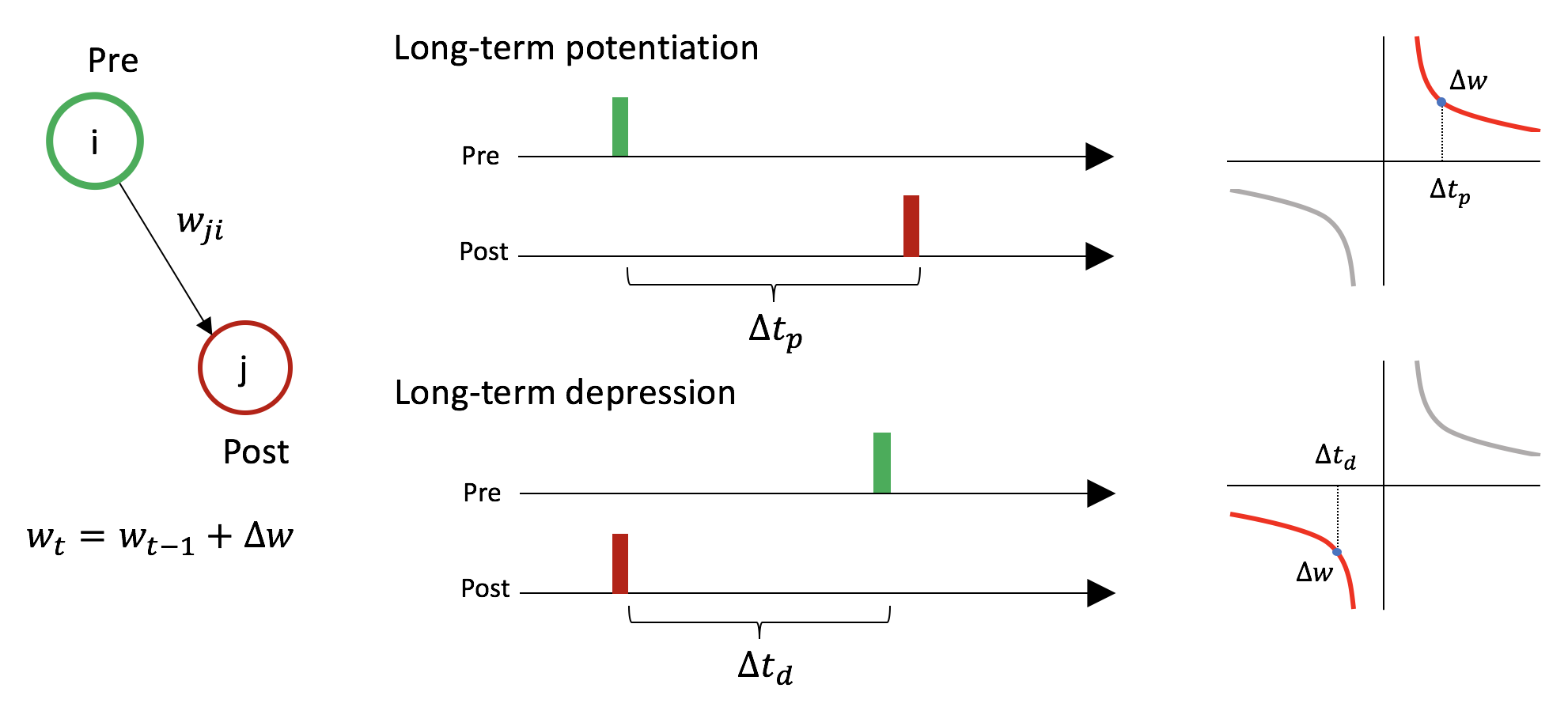}
\caption{Symmetric spike-timing dependent plasticity between two neurons. Let us consider presynaptic neuron $i$, postsynaptic neuron $j$, and the connection $w_{i,j}$ from $i$ to $j$. If neuron $i$ fires just before neuron $j$ (i.e., $\Delta t_p< 20$~ms), the synaptic weight increases by $\Delta w$. If neuron $i$ fires just after neuron $j$ (i.e., $\Delta t_d<20$~ms), the synaptic weight decreases by $\Delta w$.}
\label{fig:stdp_dynamics}
\end{figure}

STDP causes changes in synaptic weights between two firing neurons depending on the timing of their activity~(Fig.~\ref{fig:stdp_dynamics}).
For a presynaptic neuron $i$, postsynaptic neuron $j$, and the connection $w_{i,j}$ from $i$ to $j$: if neuron~$i$ fires just before neuron~$j$, $w_{i,j}$ increases (long-term potentiation [LTP]), and if neuron~$i$ fires just after neuron~$j$, $w_{i,j}$ decreases (long-term depression [LTD]).

\begin{figure}[t]
\centering
\includegraphics[width=15cm]{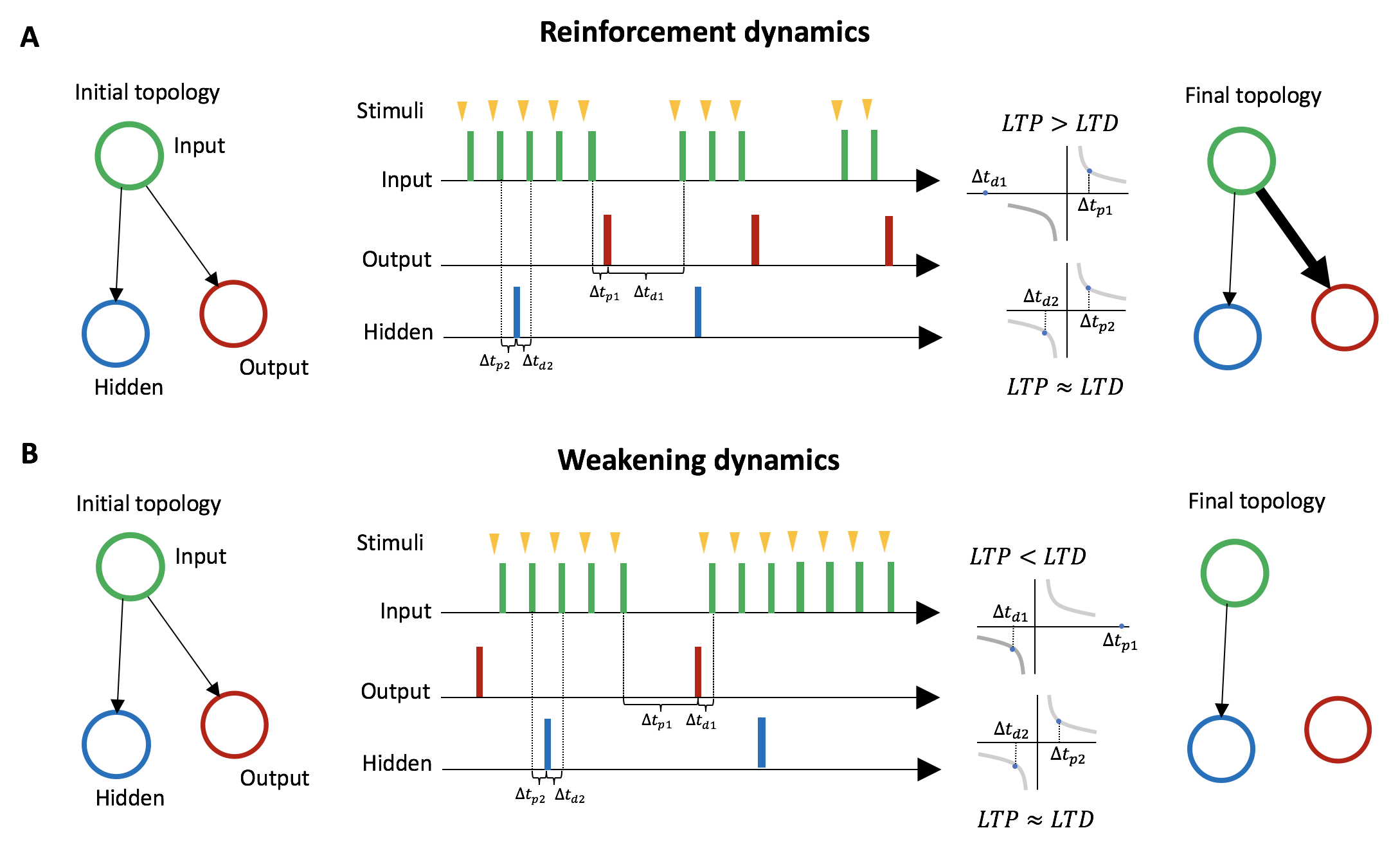}
\caption{Learning by Stimulation Avoidance in three neurons: input neuron, output neuron, and hidden neuron. A: Reinforcement dynamics of LSA. For an embodiment where if the output neuron fires right after stimulation from its environment, the stimulation is temporarily removed, the time window $\Delta t_{p1}$ between spikes of the input neuron and spikes of the output neuron gradually becomes smaller on average than the time window $\Delta t_{d1}$ between spikes of the output neuron and spikes of the input neuron. LTP being stronger than LTD, the connection weight from the input neuron to the output neuron increases. On the other hand, the connection from the input neuron to the hidden neuron barely changes, as the time windows $\Delta t_{p2}$  and $\Delta t_{d2}$ are similar on average. B: Weakening dynamics of LSA. For an embodiment where if the output neuron fires, stimulation to the input neuron starts, the time window $\Delta t_{p1}$ between spikes of the input neuron and spikes of the output neuron gradually becomes larger on average than the time window $\Delta t_{d1}$ between spikes of the output neuron and spikes of the input neuron. The effect of LTD become stronger than the effect of LTP, and the connection weight from the input neuron to the output neuron decreases. On the other hand, the connection from the input neuron to the hidden neuron barely changes.}
\label{fig:lsa_dynamics}
\end{figure}

In LSA, two mechanisms for avoiding stimulation emerge based on STDP. We explain these two mechanisms through a minimal case with three neurons: an input neuron, an output neuron, a hidden neuron. The input neuron is connected to the two other neurons~(Fig.~\ref{fig:lsa_dynamics}). The hidden neuron has no effect on other neurons or on the environment, and fires randomly.
The first mechanism, mediated by LTP, reinforces behaviors that lead to a decrease in stimulation~(Fig.~\ref{fig:lsa_dynamics}-A). We assume an embodiment in which if the output neuron fires, the stimulation to the input neuron is temporarily and immediately removed (the action of firing leads to a decrease of external stimulation). This leads the connection weight from the input neuron to the output neuron to increase, because on average the effect of LTP is stronger than the effect of LTD. On the other hand, the connection from the input neuron to the hidden neuron barely changes, because on average the effect of LTP and LTD are similar. Thus behaviors leading to a decrease in stimulation are reinforced. 
The second mechanism, mediated by LTD, is the weakening of behaviors leading to an increase in stimulation~(Fig.~\ref{fig:lsa_dynamics}-B). We assume an embodiment in which if the output neuron fires, then stimulation from the environment to the input neuron starts (the action of firing leads an increases of external stimulation). In that case, the connection weight from the input neuron to the output neuron decreases because on average the effect of LTD is stronger than the effect of LTP. The connection from the input neuron to the hidden neuron barely changes because the effects of LTP and LTD are equivalent on average. Thus behaviors leading to increases in stimulation are weakened. 

We explained LSA in a minimal case for the sake of clarity, but these dynamics work in larger networks that can express a greater variety output patterns, as we demonstrated in~\cite{Sinapayen2016, Masumori2017, Masumori2018D}. Among all output patterns, output patterns leading to a decrease in stimulation are reinforced by the first mechanism; output patterns leading to an increase in stimulation are weakened by the second mechanism. 
There is one limitation to the scalability of these networks: the bigger the network, the more internal noise the output neurons receive, and the harder it is for the network to learn a task. We previously showed that there is a lower limit of signal-to-noise ratio (SNR) in the network for LSA. The SNR decreases when the network size increases, and at the size of 60,000 neurons the network cannot learn even simple behaviors (\cite{Masumori2018D}).
Below, we discuss the conditions for the emergence of various behaviors in networks subject to LSA.


We first demonstrate the existence of three conditions required to obtain reactive behavior in biological networks as well as simulated networks (Section~\ref{sec:reaction}). We then show that a few modifications in the topology allow simulated networks to learn to predict external stimuli (Section~\ref{sec:prediction}). Finally, we explain how the reactive and predictive structures can be coupled to produce proactive behavior (Section~\ref{sec:proactive}) before discussing how inductive behavior can emerge from predictions (Section~\ref{sec:inductive}).

\section{Reactive behavior in biological and simulated networks}
\label{sec:reaction}

Let us remind the definition of reactive agents from the introduction of this paper:

\textbf{Definition:} Reactive agents learn during their lifetime how to react to environmental stimuli.

Even without neural plasticity, an agent can act reactively using hard-wired abilities as Darwinian creatures do. However, hard-wired reactive behavior can have negative consequences if environmental changes happen during an individual life time. For example, a behavior resulting in a food reward might result in getting poisoned in the future. In an environment that changes rapidly, learning reactive behavior is an effective way to help the survival of the agent. In this section, we focus on some necessary conditions to learn reactive behavior.

\begin{figure}[t]
\begin{center}
\includegraphics[width=10cm]{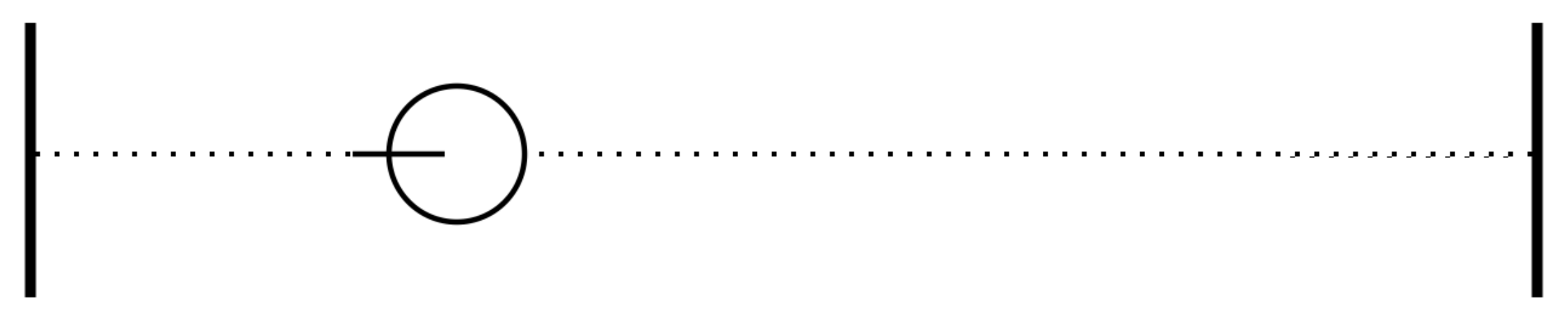}
\end{center}
\caption{A robot is moving on a line. At the end of the line, bumping into a wall causes stimulation. Turning around stops the stimulation by allowing the robot to move away from the wall.}
\label{fig:robot_1d}
\end{figure}

In a previous study (\cite{Sinapayen2016}), we showed that spiking networks with STDP exhibit LSA.
The behaviors learned by these networks is reactive: they learn to perform an action after receiving a certain type of stimulation.
We offer a simple example in Fig.~\ref{fig:robot_1d}. A robot is moving on a line. At the end of the line, bumping into a wall causes stimulation through distance sensors. Turning around stops the stimulation by allowing the robot to move away from the wall. The robot gradually learns to turn away when its distance sensors are stimulated by the walls.

\begin{figure}[t]
\begin{center}
\includegraphics[width=12cm]{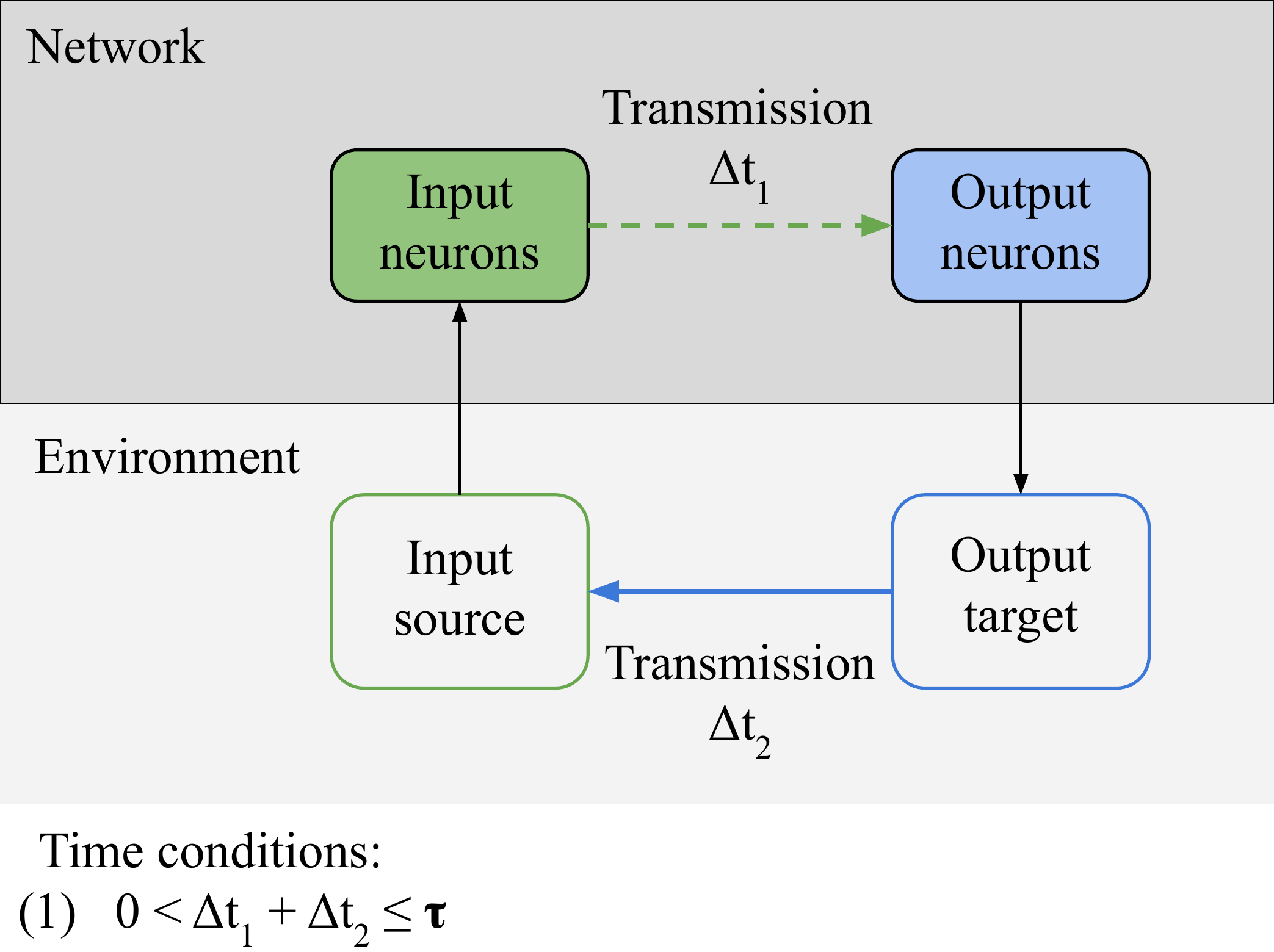}
\end{center}
\caption{Necessary conditions for reactive behavior: \textbf{Connectivity} condition for the network: relevant information from the environment must be able to reach the actuators of the agent. In the specific case of LSA, it means that input neurons must be able to directly or indirectly transmit stimulation from the environment to the output neurons. \textbf{Controllability} condition for the environment: there must be an output pattern from the network that can inhibit the stimulation through some action. Finally, there is a time condition: the input-output loop must be closed in less time than a specific time window $\tau$. The value of $\tau$ depends on the memory of the network. It is the time window during which the network can evaluate the consequences of a specific action.}\label{fig:conditions_reactive}
\end{figure}

In this section, we focus on the necessary conditions for reactive behavior to be learned (Fig.~\ref{fig:conditions_reactive}).
We identify one qualitative condition for the network, \textbf{Connectivity: relevant information from the sensors must be able to reach the actuators of the agent.} In the specific case of LSA, it means that input neurons must be able to directly or indirectly transmit stimulation from the environment to the output neurons. This condition can be broken if intrinsic noise is destroying the signal or if the path from sensors to actuators is destroyed. This condition is simple, but we build on it in following sections.
There is another qualitative condition for the environment, \textbf{Controllability: there exists a subset of outputs from the agent that can modify the source of the input in the environment.} This definition is a special case of the definition of control by Klyubin et al (\cite{Klyubin2005}). In the specific case of LSA, it means that there is an output pattern from the network that can inhibit the stimulation through an action (e.g. turning away from the wall); it can also mean that there is an action from the network that can start the stimulation to the input neurons, in which case this action will be avoided by the network in the future.
The last condition is a quantitative time constraint linking the network and the environment: \textbf{(1) the input-output loop must be closed in less time than $\tau$.} $\tau$ is the time window during which the network can evaluate the consequences of a specific action that it took. This time constant depends on the memory of the network. For example, in a simulated minimal spiking network with two neurons and in the absence of long term memory, $\tau$ is equal to 20~ms (the time constant of the STDP learning rule). The weights of a neuron's connections are only changed by STDP during this 20~ms time window before or after the neuron spikes, so an action by the network in response to a stimulation must take effect in the environment in less than 20~ms, for the action to be associated to the input and learned by the network.

A reactive agent must respect these conditions. In the following subsections, we explore the consequences of these conditions and show that they are necessary for the network to learn reactive behavior based on results from our previous studies.

\subsection{Controllability}

In a previous paper, we showed that simulated networks and biological networks can learn the wall avoidance task in a one-dimensional environment (Fig~\ref{fig:robot_1d}; \cite{Masumori2018}) through LSA. This is reactive behavior, as the robot must react to the wall by turning around. In those studies we compared the results when Controllability is respected (the target output from the network stops the stimulation immediately, $\Delta t_2$~=~0~ms) and when Controllability is not respected (the stimulation is random and no output can stop it). The Connectivity condition was respected in both conditions.
In the controllable setting, the networks learned to react to the stimulation by firing the expected output; in the uncontrollable setting, the networks did not learn to react to the input. Controllability is therefore necessary for proper learning in both biological and simulated networks.

\subsection{Connectivity}

In a previous study, we also evaluated the relation between connectivity and learning success in biological networks (\cite{Masumori2018D}). To evaluate the connectivity between the input neurons and the output neurons, we defined the connectivity measure as the ratio of connections with low time delay between input neurons and output neurons. We defined a success measure as the decrease of reaction time (time between reaching the wall and turning away from it). We found a strong correlation between connectivity and success; in addition, if the connectivity measure is zero (no appropriate connections between input and output), the network cannot learn the behavior to avoid the stimulation. Connectivity is therefore necessary for proper learning in biological networks, and although we have not yet conducted simulation experiments, we argue that this condition should be respected in simulated networks. 

Since, for LSA, the loop formed by stimuli - input neurons - output neurons - feedback from output to stimuli should be closed within a specific time window, it is clear that a time condition is required.
In the previous study, we showed that the this loop must be closed in 40 ms in simulated networks with 100 neurons (\cite{Masumori2018D}). The embodiment was idealized: the time delay for executing an action was dismissed. However, in nature there are large differences between the timescale of synaptic plasticity and timescale of behavior: e.g., in Drosophila, synaptic plasticity lasts a few milliseconds, behaviors last seconds(\cite{Drew2006}). This difference becomes larger if for bigger and more complex bodies. One way of bridging this gap is to sustain the response to stimuli~(\cite{Drew2006}). This mechanism might be required if the embodiment is more complex.

\section{Proactive behavior in simulated networks}

    \textbf{Definition:} Proactive agents perform actions based on prediction.

Although it is difficult to discuss the evolution of prediction separately from action, we first focus on the necessary conditions for a network to learn to predict environmental input without any actions. We then add actions back into the picture and discuss proactive agents.

\subsection{Predictions}
\label{sec:prediction}

In the introduction we defined a prediction as information generated inside an agent, equal to the content of external input, but preceding that input in time. To make predictions, an agent therefore needs an internal generative model and a way to compare the output of that model to the input from the environment.

\begin{figure}[t]
\begin{center}
\includegraphics[width=12cm]{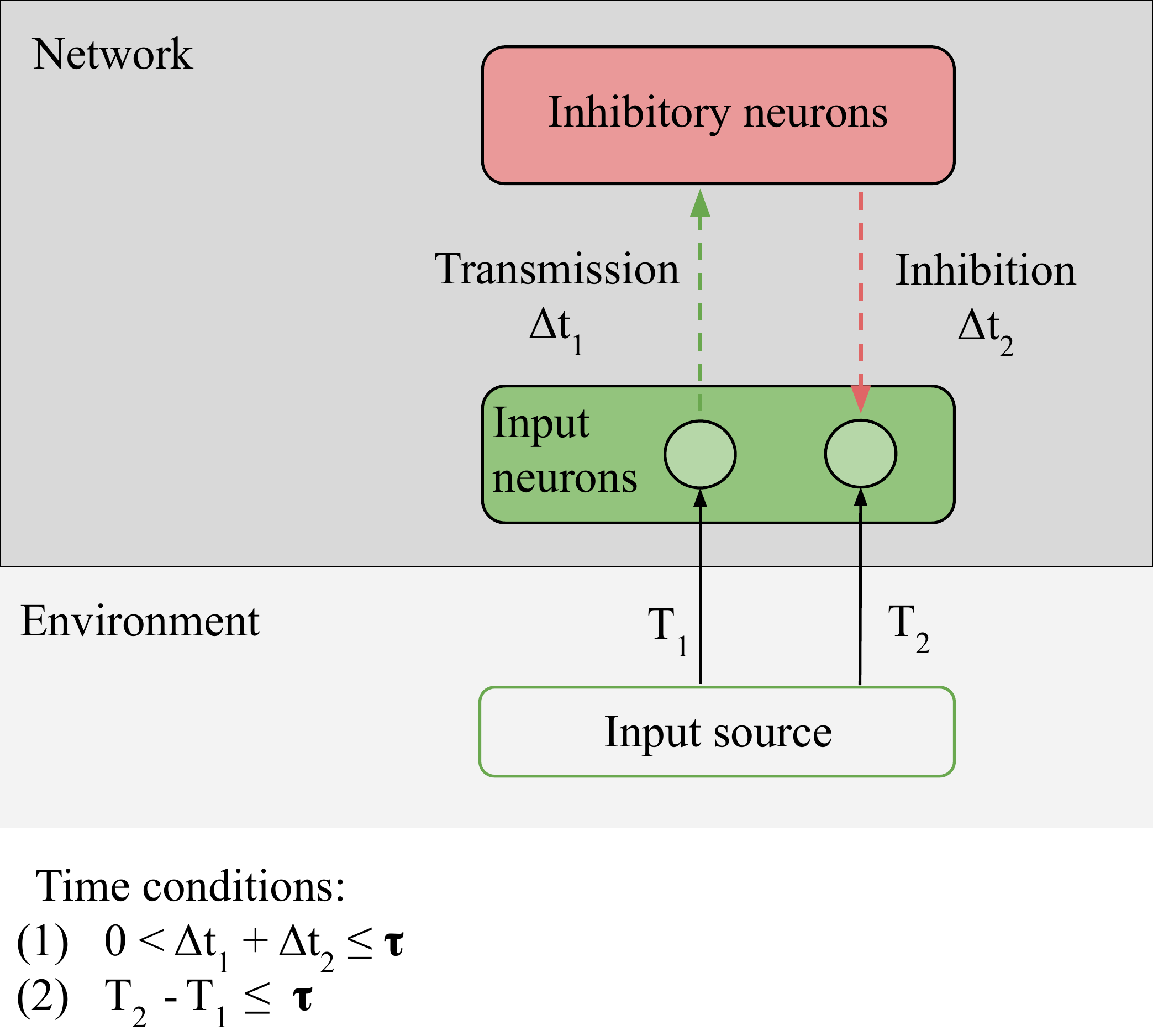}
\end{center}
\caption{Necessary conditions for prediction: \textbf{Predictability} condition for the environment: the time correlation between the stimulus at $T_1$ (unpredictable anticipatory stimulus) and the stimulus at $T_2$ (predictable target stimulus) must be reliable, i.e. the environment must provide predictable stimuli in order for predictions to be learned.
\textbf{Connectivity} condition for the network: input neurons must be able to directly or indirectly transmit and receive stimulation from the inhibitory neurons. Time conditions: (1) the transmission and the inhibition must be closed in less time than $\tau$; (2) time condition: the interval between the anticipatory stimulus and the target stimulus must be closed in less time than $\tau$. The value of $\tau$ depends on the memory of the network. 
}
\label{fig:conditions_prediction}
\end{figure}

In Fig.~\ref{fig:conditions_prediction}, we hypothesize that the comparison operation is done by inhibitory neurons~(\cite{Buonomano2000, Rao2001, Wacongne2012}). These neurons can compute prediction errors by inhibiting external stimulation: the error is null if the output of the inhibitory neurons and the external stimulation are exactly opposite.
Since the STDP rule does not change, in the case of LSA the only evolutionary step between reactive and proactive agents is the addition of inhibitory neurons in the network.

On the network, we still have the \textbf{Connectivity} condition: relevant information from the sensors must be able to reach the comparison units of the network. In the particular case of LSA, the prediction signal and the input to be predicted come to the inhibitory neurons; Connectivity is respected if the input neurons are able to directly or indirectly transmit and receive stimulation from the inhibitory neurons. This is the condition that allows us to consider predictions as information generated by the network, equal to the information of the target input, but preceding it: the inhibitory neurons must fire just before the input neurons in order to suppress the incoming stimulation.
On the environment, we have one new condition, \textbf{Predictability}: the time correlation between the stimulus at $T_1$ (unpredictable anticipatory stimulus) and the stimulus at $T_2$ (predictable target stimulus) must be reliable, i.e. the environment must provide predictable stimuli in order for predictions to be learned. In the case of LSA, the condition is strict: $T_1$ can be random, but $T_1-T_2\approx$~constant, or the prediction cannot be learned.
Not all environments respect this condition: for example, at micro-scales, the motion of small particles is stochastic.
 
There are also two time conditions; (1) is unchanged, the input-output loop must be closed in less time than $\tau$. The new condition is \textbf{(2) The time delay between the two stimuli must be smaller than the total processing time of the network}. This condition simply states that the network cannot generate predictions on a bigger timescale than the timescale of its own memory.

Our previous results with a simulated minimal network consisting of three neurons (1 anticipatory input neuron stimulated at $T_1$, 1 target input neuron stimulated at $T_2$ and 1 inhibitory neuron in between), which satisfies these conditions, could predict a simple causal sequence of stimuli (\cite{Masumori2018D}). Our preliminary results show that if the time interval between $T_1$ and $T_2$ becomes large, the network is not able to learn to predict the sequence.

This suggests that our proposed topology makes the network strengthen the path from anticipatory to target neurons, and that the Predictability condition and the time condition (2) are necessary to learn predictions. 

Therefore predictive abilities can evolve from a reactive agent by adding only one element to its neural network: inhibitory connections.
In the next subsection, we discuss the necessary conditions to obtain a agent that not only predicts inputs, but also acts on these predictions.

\subsection{Proactive behavior}
\label{sec:proactive}

The reactive agent discussed in Section~\ref{sec:reaction} can only initiate an action in relation to a stimulus after starting to receive that stimulus. In the worst case, even if the agent learns a reaction to a damaging stimulus, it cannot avoid the damage itself. If only the agent could predict the damage when getting the anticipatory stimulus, it could initiate an avoiding behavior before getting damaged. 
In this purely speculative section, we discuss how prediction and action can be combined into proactive behavior.

\begin{figure}[t]
\begin{center}
\includegraphics[width=12cm]{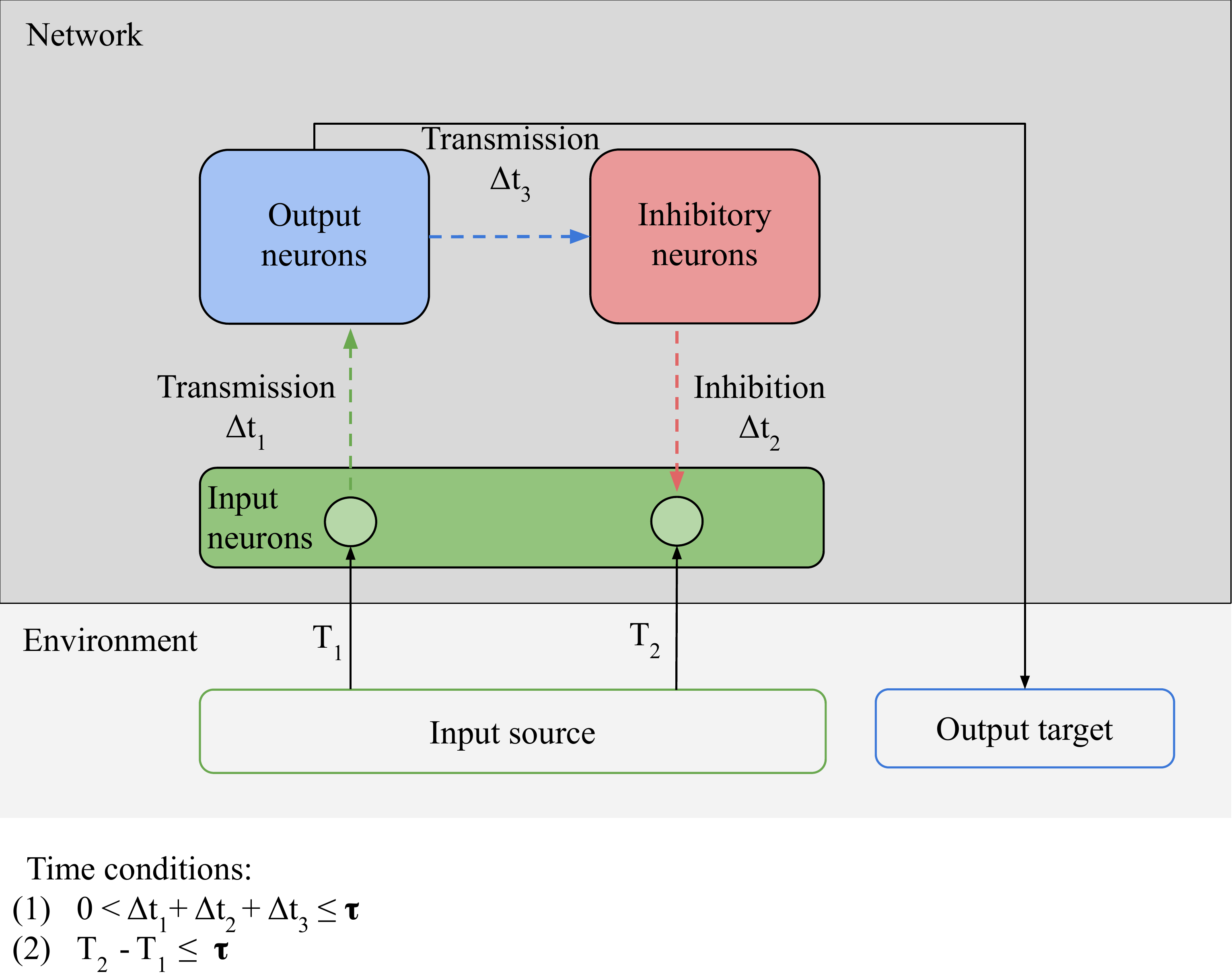}
\end{center}
\caption{Necessary conditions for proactive behavior: \textbf{Predictability} condition for the environment: the time correlation between the stimulus at $T_1$ (unpredictable anticipatory stimulus) and the stimulus at $T_2$ (predictable target stimulus) must be reliable, i.e. the environment must provide predictable stimuli in order for predictions to be learned.
\textbf{Connectivity} condition for the network: input neurons must be able to transmit stimulation to inhibitory neurons via output neurons and receive stimulation from the inhibitory neurons.
Time conditions: (1) the interval between the anticipatory stimulus and the target stimulus must be closed in less time than $\tau$; (2) the time window between transmission and inhibition must be smaller than $\tau$. The value of $\tau$ depends on the memory of the network. }
\label{fig:conditions_proactive}
\end{figure}

In Fig.~\ref{fig:conditions_prediction}, there is only minimal processing happening between the reception of an input and the next prediction. Fig.~\ref{fig:conditions_proactive} shows how to leverage more complex processing. For clarity, the output and input neurons are separated, but in the simplest case a neuron can act both as an input (by receiving external stimulation) and as an output (by outputting directly to an inhibitory neuron).
Here the task of the output neurons is to activate the right inhibitory neurons at the right time: the output neurons play the role of generative model. After valid predictions start being learned by the network, these predictions and the model that generates them can be harnessed not simply by the inhibitory neurons, but can be used to guide the behavior of the agent. Now the agent respects our definition of proactive behavior: cued by an anticipatory stimulus, it generates a prediction, and acts based on that prediction. For example, learning to move to avoid damage following an anticipatory stimulus.  

From an evolutionary point of view, predictions can only evolve if they provide increased fitness to the agent. The proactive agent must therefore either evolve directly from a reactive agent, or predictions must have evolved as a side effect of some other fitness-increasing process. How is this possible? A plausible evolutionary path might have looked like this: first, reactive agents with increased number of neurons are favored by evolution thanks to their bigger repertoire of reactive behaviors. Since coupling several excitatory neurons together leads to over-excitation of the network (maladaptive, synchronized bursts in vitro (\cite{Wagenaar2005})), mutations leading to the apparition of inhibitory neurons are favored. Even randomly coupled, inhibitory neurons tend to stabilize spiking networks (\cite{Brunel2000}). In this random structure, some of the inhibitory neurons will learn predictions because of LSA, even if the predictions are not used by the agent. Eventually, the agents that randomly learn to make use of these predictions are rewarded by higher fitness, and the structure of the network becomes less random and closer to our proposed structure, to favor the production of predictions. We could then have proactive agents.

Previous studies have shown that spiking neural networks with reward-modulated synaptic plasticity can learn proactive behaviors, as in reinforcement learning (\cite{Vasilaki2009}). These models require neuromodulators such as dopamine. We argue that these neuromodulators are not required for proactive behavior, i.e. proactive agents could well evolve without dopaminergic neurons. 


\section{Speculating About Inductive Agents}
\label{sec:inductive}

\textbf{Definition:} Inductive agents make generalizations about past stimuli, and react to new stimuli based on these generalizations.

The proactive agent can learn temporal sequences of stimuli, but it cannot extract relevant features to generalize its predictions. It is unable to judge the similarity between two stimuli, and must learn predictions anew for every single stimulus.

The last step of the path is the inductive agent. Its ability to perform classification is an advantage when the environment is variable or noisy. In these conditions, the agent must learn to abstract relevant signals from variable inputs.

Here the ``similarity" of inputs can be defined in relation to the predictions that they elicit. For example, if a set of inputs A' are all just noisy versions of input A, they should lead to the same predictions and can therefore be considered similar; A' and A belong to the same class. ``Noise" is one type of variation, but there can be others that still lead to ``similar" predictions, where this time prediction similarity is defined relatively to the actions afforded by the predictions. The notion of similarity can in this way be propagated from the bottom up through all 3 functions: action, prediction, and classification. Same action or same prediction caused by some inputs means that the inputs belong to the same class.

With artificial networks, classification is typically considered in the context of labelled data. The labels are used to compute an error signal that is propagated from the output neurons to the input neurons. For an agent in the biological world, there are usually no labels, and reward/error signals from the environment are too sparse to learn to classify even a few thousands of inputs.

Predictions can provide the abundant error signal necessary to learn classification: each time step provides a prediction error. The connections in the network can be optimized at each time step to give better predictions. Let us suppose a cost on updating the connections in the network: changing the weight of one connection is less costly than changing the weight of two connections, etc. We now have the perfect setup for the emergence of hierarchies of classes. 

In hierarchical networks, the input from lower layers is aggregated in upper layers into classes that are more and more general. The most invariant properties of the input end up being represented by the classes at the top of the hierarchy.
In our proposal, the neurons receiving the raw prediction error are the neurons close to the input neurons, lower in the hierarchy (note that this is the opposite of Deep Learning Neural Network architectures, where the error signal is propagated from higher layers to lower input layers).
The lower layers can remove as much variance as possible from the input before passing it to upper neurons. Most of the prediction error, due to the most variable properties of the input, will be corrected in these lower layers. The remaining error is progressively corrected by updating weights in upper layers. Our proposal therefore minimizes the cost of learning by having the entry point of the error signal close to the input neurons.

\cite{foldiak1991learning} demonstrate a similar result in simulation: with a local learning rule (minimizing the number of weights being updated), a predictive neural network learns invariances in temporal sequences and becomes able to do simple generalizations through groups of cells reacting to similar inputs.

Thus we hypothesize that predictive abilities are a necessary condition to obtain the abundant error signal required to learn classification, and updating costs are a necessary condition to obtain the hierarchical, modular structure characteristic of generalization.
A consequence of this hypothesis is that classification emerges as a way to improve the quality of predictions.

\section{Discussion}
\label{sec:discussion}

In this paper, we have outlined necessary, practical evolutionary steps to go from a reactive to an inductive agent. The steps require 4 conditions to be satisfied:

\begin{itemize}
    \item  \textbf{Connectivity}: relevant information from the sensors must be able to reach the comparison units of the network
    \item \textbf{Predictability}: the time correlation between unpredictable anticipatory stimuli and predictable target stimuli must be reliable
    \item \textbf{Time condition (1)}: the input-output loop must be closed in less time than a specific time window $\tau$
    \item \textbf{Time condition (2)}:  the transmission and the inhibition must be closed in less time than $\tau$
\end{itemize}

The reactive agent, although requiring preliminary conditions (Controllability, Connectivity, and time condition~(1)), does not require a well-designed network structure. The network requires only excitatory neurons for the agent to learn reactive behaviors. Prediction requires two additional conditions: Predictability and time condition~(2). A network respecting simple structural rules and containing inhibitory neurons can learn predictive behaviors. The proactive agent, in addition to the four previous conditions, requires a more well-designed structure. Our results support the hypothesis that in evolutionary history, at first, reactive agents emerge with simple structured network with limited type of neurons (excitatory neurons) and neuronal plasticity. Proactive agents evolve from reactive agents with more structured network with various type of neurons (excitatory neurons and inhibitory neurons).
Inductive agents, in addition to the previous conditions, require to have a cost on updating the weights of the network connections.

Our proposed action-prediction-classification evolutionary path therefore requires, in this order, the evolution of: excitatory neurons, inhibitory neurons, cost-based network structure.

What experiments could consolidate or disprove our hypotheses?
To demonstrate our proposed step for the transition from action to prediction, we must show that inhibitory neurons are used for prediction. This could be demonstrated by finding a positive correlation between the activity of inhibitory neurons and the value of target stimuli in the environment, especially in animals with a simple nervous system.
 
The transition from prediction to classification might be supported by extending the experiments of Kwong et al. (\cite{kwon2008internal}). They showed that generalization might have evolved from networks with predictable internal dynamics. If "predictable internal dynamics" reflect the fact that the networks have learned to predict the environment, and this environmental predictablity is mirrored by internal predictability, then this would support our hypothesis of a transition from predictive networks to generalizing networks. In addition, if the predictable networks show a hierarchical internal structure with error correction happening primarily at lower layers, this would further support our proposed mechanism for the transition from prediction to classification.
This would demonstrate the relationship between classification and prediction.

Our four types of agents depart from Dennett's four classes in two major ways. First, we are only interested in learning happening during the lifetime of an agent: Dennett's Darwinian creatures (which have no learning ability) and Gregorian creatures (which pass knowledge on beyond the timescale of a lifetime) are out of the scope of our considerations.
Secondly, we introduce the inductive agent as a step between Popperian and Gregorian creatures, thus arguing that the ability to generalize through the classification of stimuli is evolutionary distinct from predictive abilities and different from trans-generational learning. Focusing on the evolutionary links between agents' abilities made this new classification necessary.

There is recent interest in common AI approaches towards predictive networks, yet the connection between action, prediction, and classification is rarely explored. Indeed, until recently, disembodied classifying networks represented most of the state of the art. From there, interest is slowly shifting towards predictive networks. Our results suggest that the opposite direction, focusing on prediction and from there evolving classification abilities, can be a fruitful area of research.

\section*{Conflict of Interest Statement}

The authors declare that the research was conducted in the absence of any commercial or financial relationships that could be construed as a potential conflict of interest.

\section*{Acknowledgements}

We thank all the reviewers for their thoughtful comments that helped us improve this paper.

\section*{Author Contributions}

AM and LS contributed equally to this paper as first authors, writing the drafts and revising the manuscript. IT contributed conception and design of the study. All authors contributed to manuscript revision, read and approved the submitted version.

\section*{Funding}
AM is supported by Grant-in-Aid for JSPS Fellows (16J09357). 
This work is partially supported by MEXT project ``Studying a Brain Model based on Self-Simulation and Homeostasi'' in
Grant-in-Aid for Scientific Research on Innovative Areas
``Correspondence and Fusion of Artificial Intelligence and Brain Science''
(19H04979)

\bibliographystyle{frontiersinSCNS_ENG_HUMS} 
\bibliography{frontiers}

\begin{thebibliography}{24}
\providecommand{\natexlab}[1]{#1}
\expandafter\ifx\csname urlstyle\endcsname\relax
  \providecommand{\doi}[1]{doi:\discretionary{}{}{}#1}\else
  \providecommand{\doi}{doi:\discretionary{}{}{}\begingroup
  \urlstyle{rm}\Url}\fi
\providecommand{\selectlanguage}[1]{\relax}
\providecommand{\bibAnnoteFile}[1]{%
  \IfFileExists{#1}{\begin{quotation}\noindent\textsc{Key:} #1\\
  \textsc{Annotation:}\ \input{#1}\end{quotation}}{}}
\providecommand{\bibAnnote}[2]{%
  \begin{quotation}\noindent\textsc{Key:} #1\\
  \textsc{Annotation:}\ #2\end{quotation}}

\bibitem[{Brunel(2000)}]{Brunel2000}
Brunel, N. (2000).
\newblock {Dynamics of sparsely connected networks of excitatory and inhibitory
  spiking neurons}.
\newblock \emph{Journal of computational neuroscience} 8, 183--208
\bibAnnoteFile{Brunel2000}

\bibitem[{Buonomano(2000)}]{Buonomano2000}
Buonomano, D.~V. (2000).
\newblock Decoding temporal information: A model based on short-term synaptic
  plasticity.
\newblock \emph{Journal of Neuroscience} 20, 1129--1141.
\newblock \doi{10.1523/JNEUROSCI.20-03-01129.2000}
\bibAnnoteFile{Buonomano2000}

\bibitem[{Caporale and Dan(2008)}]{Caporale2008}
Caporale, N. and Dan, Y. (2008).
\newblock {Spike timing-dependent plasticity: a Hebbian learning rule.}
\newblock \emph{Annual review of neuroscience} 31, 25--46.
\newblock \doi{10.1146/annurev.neuro.31.060407.125639}
\bibAnnoteFile{Caporale2008}

\bibitem[{Chung et~al.(2009)Chung, Kwon, and Choe}]{chung2009evolution}
Chung, J.~R., Kwon, J., and Choe, Y. (2009).
\newblock Evolution of recollection and prediction in neural networks.
\newblock In \emph{2009 International Joint Conference on Neural Networks}
  (IEEE), 571--577
\bibAnnoteFile{chung2009evolution}

\bibitem[{Dennett(1995)}]{Dennett1995}
Dennett, D.~C. (1995).
\newblock \emph{{Darwin's Dangerous Idea: Evolution and the Meanings of Life}}
  (New York: Simon {\&} Schuster)
\bibAnnoteFile{Dennett1995}

\bibitem[{Drew and Abbott(2006)}]{Drew2006}
Drew, P.~J. and Abbott, L.~F. (2006).
\newblock Extending the effects of spike-timing-dependent plasticity to
  behavioral timescales.
\newblock \emph{Proceedings of the National Academy of Sciences} 103,
  8876--8881.
\newblock \doi{10.1073/pnas.0600676103}
\bibAnnoteFile{Drew2006}

\bibitem[{Edwards et~al.(2017)Edwards, Vetter, McGruer, Petro, and
  Muckli}]{Edwards2017}
Edwards, G., Vetter, P., McGruer, F., Petro, L.~S., and Muckli, L. (2017).
\newblock Predictive feedback to v1 dynamically updates with sensory input.
\newblock \emph{Scientific Reports} 7, 16538.
\newblock \doi{10.1038/s41598-017-16093-y}
\bibAnnoteFile{Edwards2017}

\bibitem[{F{\"o}ldi{\'a}k(1991)}]{foldiak1991learning}
F{\"o}ldi{\'a}k, P. (1991).
\newblock Learning invariance from transformation sequences.
\newblock \emph{Neural Computation} 3, 194--200
\bibAnnoteFile{foldiak1991learning}

\bibitem[{{Klyubin} et~al.(2005){Klyubin}, {Polani}, and
  {Nehaniv}}]{Klyubin2005}
{Klyubin}, A.~S., {Polani}, D., and {Nehaniv}, C.~L. (2005).
\newblock Empowerment: a universal agent-centric measure of control.
\newblock In \emph{2005 IEEE Congress on Evolutionary Computation}. vol.~1,
  128--135 Vol.1.
\newblock \doi{10.1109/CEC.2005.1554676}
\bibAnnoteFile{Klyubin2005}

\bibitem[{Kwon and Choe(2008)}]{kwon2008internal}
Kwon, J. and Choe, Y. (2008).
\newblock Internal state predictability as an evolutionary precursor of
  self-awareness and agency.
\newblock In \emph{2008 7th IEEE International Conference on Development and
  Learning} (IEEE), 109--114
\bibAnnoteFile{kwon2008internal}

\bibitem[{Masumori(2019)}]{Masumori2018D}
Masumori, A. (2019).
\newblock Homeostasis by action, prediction, selection in embodied neural
  networks (doctoral dissertation) [in press]
\bibAnnoteFile{Masumori2018D}

\bibitem[{Masumori et~al.(2017)Masumori, Sinapayen, and Ikegami}]{Masumori2017}
Masumori, A., Sinapayen, L., and Ikegami, T. (2017).
\newblock {Learning by stimulation avoidance scales to large neural networks}.
\newblock In \emph{Proceedings of the 14th European Conference on Artificial
  Life ECAL 2017} (Cambridge, MA: MIT Press), September, 275--282.
\newblock \doi{10.7551/ecal_a_048}
\bibAnnoteFile{Masumori2017}

\bibitem[{Masumori et~al.(2018)Masumori, Sinapayen, Maruyama, Mita, Bakkum,
  Frey et~al.}]{Masumori2018}
Masumori, A., Sinapayen, L., Maruyama, N., Mita, T., Bakkum, D., Frey, U.,
  et~al. (2018).
\newblock Autonomous regulation of self and non-self by stimulation avoidance
  in embodied neural networks.
\newblock \emph{The 2018 Conference on Artificial Life: A Hybrid of the
  European Conference on Artificial Life (ECAL) and the International
  Conference on the Synthesis and Simulation of Living Systems (ALIFE)} ,
  163--170\doi{10.1162/isal\_a\_00037}
\bibAnnoteFile{Masumori2018}

\bibitem[{Powers et~al.(2016)Powers, Kelley, and Corlett}]{Powers2016}
Powers, I., Albert~R., Kelley, M., and Corlett, P.~R. (2016).
\newblock Hallucinations as top-down effects on perception.
\newblock \emph{Biological Psychiatry: Cognitive Neuroscience and Neuroimaging}
  1, 393--400.
\newblock \doi{10.1016/j.bpsc.2016.04.003}
\bibAnnoteFile{Powers2016}

\bibitem[{Raman and Sarkar(2016)}]{Raman2016}
Raman, R. and Sarkar, S. (2016).
\newblock Predictive coding: A possible explanation of filling-in at the blind
  spot.
\newblock \emph{PLOS ONE} 11, 1--17.
\newblock \doi{10.1371/journal.pone.0151194}
\bibAnnoteFile{Raman2016}

\bibitem[{Rao and Ballard(1999)}]{Rao1999}
Rao, R. P.~N. and Ballard, D.~H. (1999).
\newblock Predictive coding in the visual cortex: a functional interpretation
  of some extra-classical receptive-field effects.
\newblock \emph{Nature Neuroscience} 2, 79 EP --
\bibAnnoteFile{Rao1999}

\bibitem[{Rao and Sejnowski(2001)}]{Rao2001}
Rao, R. P.~N. and Sejnowski, T.~J. (2001).
\newblock Spike-timing-dependent hebbian plasticity as temporal difference
  learning.
\newblock \emph{Neural Comput.} 13, 2221--2237.
\newblock \doi{10.1162/089976601750541787}
\bibAnnoteFile{Rao2001}

\bibitem[{Seth et~al.(2012)Seth, Suzuki, and Critchley}]{Seth2012}
Seth, A., Suzuki, K., and Critchley, H. (2012).
\newblock An interoceptive predictive coding model of conscious presence.
\newblock \emph{Frontiers in Psychology} 2, 395.
\newblock \doi{10.3389/fpsyg.2011.00395}
\bibAnnoteFile{Seth2012}

\bibitem[{Sinapayen et~al.(2017)Sinapayen, Masumori, and
  Ikegami}]{Sinapayen2016}
Sinapayen, L., Masumori, A., and Ikegami, T. (2017).
\newblock {Learning by stimulation avoidance: A principle to control spiking
  neural networks dynamics}.
\newblock \emph{PLOS ONE} 12, e0170388.
\newblock \doi{10.1371/journal.pone.0170388}
\bibAnnoteFile{Sinapayen2016}

\bibitem[{Suzuki et~al.(2017)Suzuki, Roseboom, Schwartzman, and
  Seth}]{Suzuki2017}
Suzuki, K., Roseboom, W., Schwartzman, D.~J., and Seth, A.~K. (2017).
\newblock A deep-dream virtual reality platform for studying altered perceptual
  phenomenology.
\newblock \emph{Scientific Reports} 7, 15982.
\newblock \doi{10.1038/s41598-017-16316-2}
\bibAnnoteFile{Suzuki2017}

\bibitem[{Vasilaki et~al.(2009)Vasilaki, Frémaux, Urbanczik, Senn, and
  Gerstner}]{Vasilaki2009}
Vasilaki, E., Frémaux, N., Urbanczik, R., Senn, W., and Gerstner, W. (2009).
\newblock Spike-based reinforcement learning in continuous state and action
  space: When policy gradient methods fail.
\newblock \emph{PLOS Computational Biology} 5, 1--17.
\newblock \doi{10.1371/journal.pcbi.1000586}
\bibAnnoteFile{Vasilaki2009}

\bibitem[{Wacongne et~al.(2012)Wacongne, Changeux, and Dehaene}]{Wacongne2012}
Wacongne, C., Changeux, J.-P., and Dehaene, S. (2012).
\newblock {A Neuronal Model of Predictive Coding Accounting for the Mismatch
  Negativity}.
\newblock \emph{Journal of Neuroscience} 32, 3665--3678.
\newblock \doi{10.1523/JNEUROSCI.5003-11.2012}
\bibAnnoteFile{Wacongne2012}

\bibitem[{Wagenaar et~al.(2005)Wagenaar, Madhavan, Pine, and
  Potter}]{Wagenaar2005}
Wagenaar, D.~A., Madhavan, R., Pine, J., and Potter, S.~M. (2005).
\newblock {Controlling bursting in cortical cultures with closed-loop
  multi-electrode stimulation}.
\newblock \emph{The Journal of neuroscience} 25, 680--688
\bibAnnoteFile{Wagenaar2005}

\bibitem[{Watanabe et~al.(2018)Watanabe, Kitaoka, Sakamoto, Yasugi, and
  Tanaka}]{Watanabe2018}
Watanabe, E., Kitaoka, A., Sakamoto, K., Yasugi, M., and Tanaka, K. (2018).
\newblock Illusory motion reproduced by deep neural networks trained for
  prediction.
\newblock \emph{Frontiers in Psychology} 9, 345.
\newblock \doi{10.3389/fpsyg.2018.00345}
\bibAnnoteFile{Watanabe2018}

\end{thebibliography}

\end{document}